\documentclass[procedia]{easychair}

\usepackage{doc}
\usepackage{makeidx}

\def\procediaConference{2016 Annual International Conference on Biologically Inspired Cognitive Architectures (BICA 2016)}

\title{Psychologically inspired planning method for smart relocation task}

\titlerunning{Psychologically inspired planning method for smart relocation task}

\author{
    Aleksandr I. Panov
\and
    Konstantin S. Yakovlev
}

\institute{
  Federal Research Center ``Computer Science and Control'' of Russian Academy of Sciences,
  Moscow, Russia\\
  \email{\{pan,yakovlev\}@isa.ru}
 }

\authorrunning{Panov and Yakovlev}

\begin{document}

\maketitle


\begin{abstract}
  Behavior planning is known to be one of the basic cognitive functions, which is essential for any cognitive architecture of any control system used in robotics. At the same time most of the widespread planning algorithms employed in those systems are developed using only approaches and models of Artificial Intelligence and don't take into account numerous results of cognitive experiments. As a result, there is a strong need for novel methods of behavior planning suitable for modern cognitive architectures aimed at robot control. One such method is presented in this work and is studied within a special class of navigation task called smart relocation task. The method is based on the hierarchical two-level model of abstraction and knowledge representation, e.g. symbolic and subsymbolic. On the symbolic level sign world model is used for knowledge representation and hierarchical planning algorithm, MAP, is utilized for planning. On the subsymbolic level the task of path planning is considered and solved as a graph search problem. Interaction between both planners is examined and inter-level interfaces and feedback loops are described. Preliminary experimental results are presented.
\end{abstract}


%
%

\section{Introduction}
\label{sect:introduction}

Behavior planning is one of the basic cognitive functions and it's effective realization is essential for all cognitive architectures used for robot control \cite{Emelyanov2016,Hanford2011,Trafton2013}. Typically planning algorithms for control systems used in robotics are developed and studied within a specific direction of Artificial Intelligence called automated (or intelligent) planning. Within this direction a great number of approaches has been developed so far \cite{DellaPenna2012,Richter2010,Hoffmann2001}. Most of these approaches rely on STRIPS formalism \cite{Fikes1971} and implement certain properties of the cognitive planning process characteristic for humans e.g. hierarchical pattern, backward planning, planning with incomplete information etc. At the same time AI planning methods don't handle all range of cognitive experimental facts (e.g. metacognitive control or heuristic change depending on a current planning state) and cannot provide ground for construction of consolidated model of human behavior planning process.

On the other hand a vast range of cognitive experimental facts is taken into account while developing cognitive architectures such as ACT-R, SOAR, Clarion etc. Architectures claiming to demonstrate behavior which is close to behavior exhibited by humans in complicated situations (for example while human-robot interaction) should contain the module which implements psychologically inspired action planning \cite{Sun2012a}. Further, the knowledge representation method employed in a cognitive architecture plays very important role \cite{Lieto2014}. The method should provide the association of symbolic level of action planning and subsymbolic representation level of recognition of external signals and conditions.

In the paper we consider the psychological theory of activity \cite{Leontyev2009} to represent the world model and to construct the model of planning process as a structurally equivalent of the corresponding psychological process \cite{Lieto2016}. Within this theory an agent's behavior is considered to be carried out in the course of the so-called activity directed by a motive (significance of a needed object). This activity is comprised out of a set of actions. Each action is aimed at achieving specific goal and consists of automated operations. Combination of operations forming an action is dependent on the observed conditions. While modeling the process of behavior planning actions and operations are formalized with some formal procedures. We assume that the separation of the available procedures to action and operation sets is dynamic and can vary during planning process. In any case we attribute the goal oriented actions to be symbolic procedures forming the symbolic level of the behavior planning and automated operations to be subsymbolic ones (subsymbolic level). Within our assumption at a certain stage of the planning process some operations can be raised to the symbolic level and contrariwise some actions can be automated.

To demonstrate the proposed approach we examine the special class of navigation tasks significant to robotics, e.g. smart relocation tasks in complex environments. Solution of these tasks leads to the necessity of examination of challenging technical and cognitive aspects of behavior e.g. applying several classes of actions (different types of elementary relocations, object manipulation, communicative actions), using different types of description levels (description of objects and spatial-temporal relations), ability to set not only individual but also collective goals. Collective version of the smart relocation task was considered in \cite{Panov2016a} and in the current paper we consider only individual actions and goals.

We use knowledge representation framework proposed in \cite{Osipov2014b,Osipov2015e} for world model description. Within this framework the grounded symbol \cite{Harnad1990} is defined by the four component structure entitled a sign --- name, image, significance and personal meaning. A sign as the basic element of world model can mediate a static object as well as a process or an action. A sign name conforms to linguistic rules of using the corresponding word in the agent group \cite{Steels2012}. Image component of a sign contains the recognition procedure which relies on specific features of mediated object or process. Significance component of a sign defines functional or other roles of the mediated entity in scripts (as defined by Shank \cite{Schank1972}) known to all agents in the group. Personal meaning of a sign defines roles in personal scripts known to the particular agent. Features used in recognition procedure of sign image are the links to either other signs in the world model (in case the actualized level of the given sign is the symbolic level) or to subsymbolic operations. In context of smart relocation task the subsymbolic operations are operations of planning some elementary relocations i.e. operations of path planning. Features themselves are combined into special structures --- prediction matrixes --- that define sufficient and necessary conditions of sign recognition and can be interpreted in terms of prototype theory or exemplars theory \cite{Machery2011}.

In this work we present algorithms for each level of behavior planning. On the symbolic level it is a sign planning algorithm (MAP). On the subsymbolic level it is a path planning algorithm. Special attention it is paid to linking both algorithms and description of feedback links between them which is realized within proposed knowledge representation. In the second part of the paper we present experiments carried out with presented algorithms.
\section{Smart relocation task}
\label{sect:smart}

Consider a 2D world populated with the goal-oriented intelligent agent and the artifacts of different kind some of which can be destroyed by the agent. The latter knows it's goal and can generate subgoals in order to achieve it. We concentrate now only on spatial properties of the goal state and think of the goal as of the dedicated area described in some way. In this case all the artefacts are considered to be obstacles, some of which can be removed (destroyed) by the agent. World description is available to the agent in two forms, e.g. symbolic and sub-symbolic levels. 

\begin{figure}[tb]
	\begin{centering}
		\includegraphics[width=\textwidth]{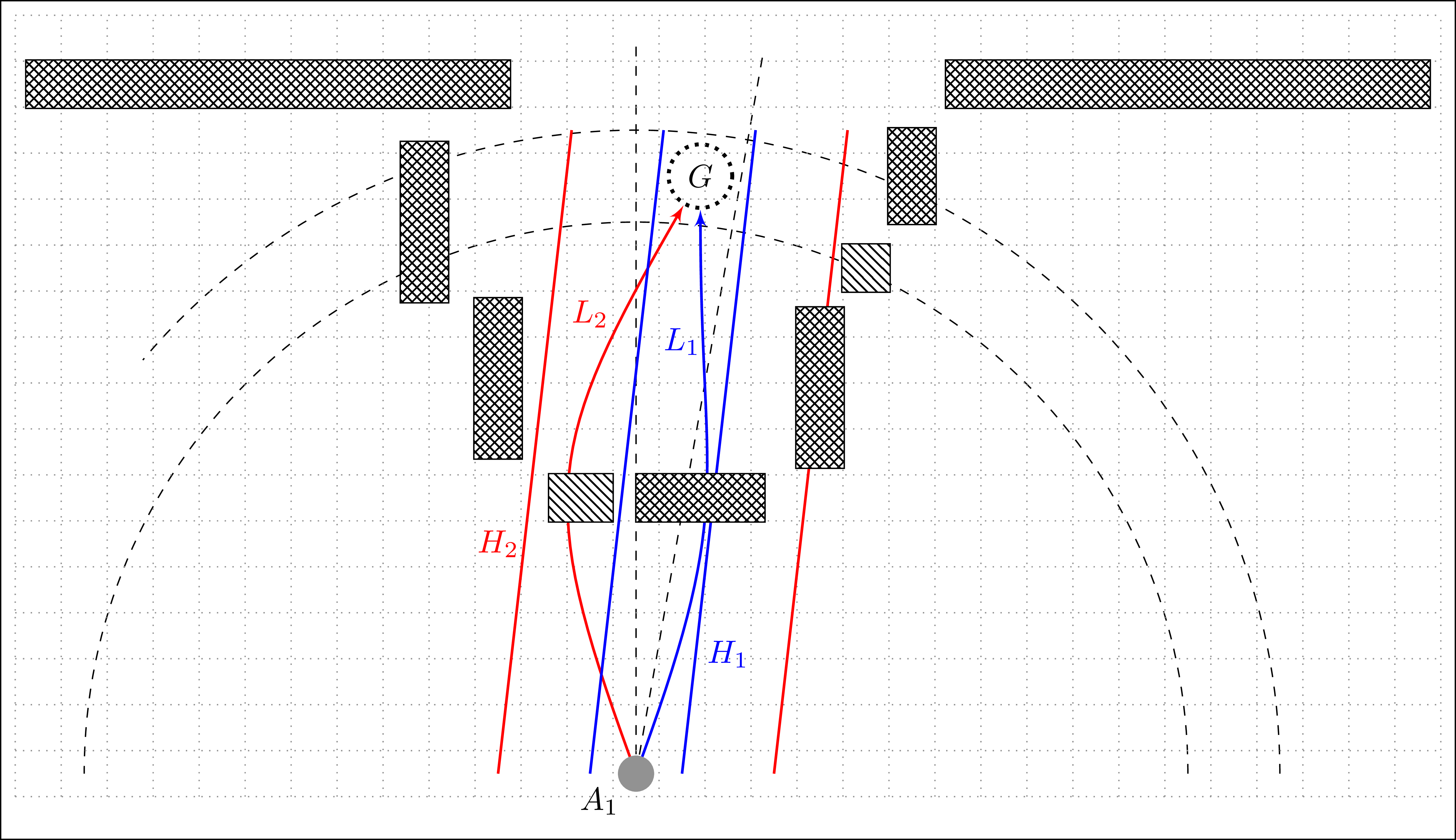}
		\caption{Single-agent smart relocation task. $H1$ and $H_2$ are permitted deviations from the straight link to the goal area. $L_1$ and $L_2$ are traces of corresponding relocation actions.}
		\label{fig:example}
	\end{centering}
\end{figure}

On the symbolic level one of the so called spatial logics can be used to describe spatial localization of agents and obstacles as these logics model human spatial reasoning in some approximation according to cognitive psychologists such \cite{Herskovits1997,Kuipers2000}. In our work fuzzy dynamic spatial logic \cite{Zadeh2012} proposed in Pospelov's situation control theory \cite{Osipov1997b} is used to describe directions and goal areas. Within this logic, the latter are specified not as precise values (45 degrees counterclockwise, 13 meters ahead) but as names of signs which mediate fuzzy directions and distances (``leftward'', ``afar''). These signs are grounded to specific values (measured in degrees and meters) via hierarchical connections between the image components of signs. In accordance with the cognitive data employed in Pospelov's research, agent-centered polar coordinate frame is used for grounding of signs corresponding to external areas: goal area, obstacle areas etc. The latter now turn into the sectors within this frame --- see Figure~\ref{fig:example}. Each agent's relocation action is represented by a sign with a name: ``moving afar in a straight line'', ``moving closer on a curve'' etc. and is grounded to a) certain fuzzy changes of distance measured from current agent location and b) certain permitted deviation from the straight connecting current location and the goal area --- see Figure~\ref{fig:example}.

On the sub-symbolic level agent's world is described as the rectangular area of 2D Euclidean space $U$: $x_{min} \leq x \leq x_{max}$, $y_{min} \leq y \leq y_{max}$. The agent is modeled as a circle of radius $R$. $U$ is comprised out of the free space $U_{free}$ and the obstacles $U_{obs}=\{obs_1,\dots obs_M\}$. Each obstacle is a polygon defined by the set of it's vertices' coordinates $obs_i=\{p_{i1}, p_{i2}, \dots, p_{ij}, \dots p_{iK_i}\}$, $p_{ij}=(x_{ij}, y_{ij})\in U$. Some obstacles can be destroyed by the agent, e.g. there exist a mapping $destr: U_{obs} \rightarrow \{true, false\}$ and if $destr(obs_i)=true$ than $obs_i$ can be destroyed. To destroy an obstacle the agent must approach it first. Goal area $G$ is considered to be a square described by the 4 pairs $(x_i, y_i)$. A path to the goal is a sequence of traversable line segments in $U_{free}$ starting from current agent's location and finishing within the goal square. We consider that a mapping exists between symbolic representation of spatial knowledge and sub-symbolic. In the simplest case this mapping is just translating the coordinates between the global Cartesian frame and the agent-centered polar frame. For example, initially, on the symbolic level goal area is characterized by 4 pairs $(r_i, \phi_i)$ defining a fuzzy sector in the agent-centered world. These coordinates can be translated to the Cartesian coordinate system and then the square which contains that sector can be estimated --- see Figure~\ref{fig:example} for details.

We now only study the case when the goal, the obstacles and the initial position of the agent form such configuration when no spatial path from start to goal initially exist. Traditional path planning methods can only state that fact but we would like to investigate the scenario further. Obviously no-path means that some obstacle is blocking it. By knowing that ``I'' can destroy some sorts of obstacles the agent can generate appropriate subgoals, approach destroyable obstacle and by removing it free it's way to the goal. We call such tasks --- smart relocation tasks (SRT). SRT can not be solved by pure path planning methods as path planers know nothing about agent's high-level actions (e.~g. destroy) and in considered case can only state that spatial path can not be constructed on a given map. A deep interaction between path planner and behavior planner is needed to successfully overcome difficulties associated with SRT solving.

\section{Method}
\label{sect:method}
\subsection{Behavior planning}
\label{sect:beh}

Semiotic network of signs (as described in introduction) $\Omega=\langle W_p,W_m,W_a\rangle$ is used for knowledge representation on the symbolic level. It consists of three semantic networks: a network on the set of sign images $W_p$, a network on the set of sign significances $W_m$ and a network on the set of sign personal meanings $W_a$. Each binary relation in a semantic network on the set of sign components is explicitly defined by the configuration of structures (prediction matrices) of these components. Each component of a sign is described with a unified model based on the neurophysiological data about neocortex columnar architecture. The structure of an image component of a sign provides a connection between symbolic and subsymbolic levels of the object or process representation in the world model.

Image component of a sign is represented by the structured set of links to another sign images or to elementary features or operations. For simplicity we call these links as features of the sign image. Features form so-called bit prediction matrices. In a prediction matrix each row corresponds to a feature and 1-bit in each column defines that this feature is required in a particular moment of time for successful recognizing an object or a process mediated by the sign. In other words the prediction matrix gives the sequence of the input feature set which should sequentially appear in the input data flow to make a decision that the sign is recognized in the flow. For example if we consider a sign with a name ``face'' then a possible variant of a prediction matrix can be the sequence of features ``nose'', ``mouth'', ``right eye'', ``left eye''.

The organization of features into such structure allows firstly to model dynamic recognition process based on saccadic eye movements. Secondly the presence of feedback links to actions that change the inner state of an agent allows to model such cognitive processes as overt and covert attention.

Image component of a sign can correspond to several prediction matrices due to the peculiarities of the learning process of sign image formation is a precedential process \cite{Skrynnik2016}. For example the prediction matrix of the ``face'' sign could be additionally presented by a feature sequence following another order: ``left eye'', ``nose'', ``mouth'', ``right eye'' or by a sequence with an additional feature: ``left eye'', ``nose'', ``mouth'', ``padded''+''right eye''. If strong sequence of columns can be extracted from all prediction matrices of the sign (e.g. there exists the feature set that always precedes another set) then the causal relationship between features of the sign is present. It can be interpreted as the sign represents a deterministic process or action. We call such signs as procedural signs (features).

Significance of a sign $s$ is a set of features (including procedural features) --- links to other signs $(s_m, s_n, \dots)$. Image component of these signs include the sign $s$ as a requirement for recognition, e.g. a prediction matrices of $s_m, s_n$ etc. contain the 1-bit in the cell corresponding to the sign $s$. It's important to note that in case of coalition behavior the set of features (significance) is preconcerted by all agents in the group. Procedural signs related to significance of another sign often are generalized and contain links to certain action roles instead links to specific objects.

Personal meanings of a sign $s$ are formed in the process of agent's activity and are instantiated agent's actions within the mediated object or instantiated relations between the sign $s$ and another sings in the certain situation. Personal meanings are associated with inner characteristics of the agent and with agent's emotional-need sphere and define any motive of agent's activity.

Behavior planning process occurring in sign world model is implemented by a MAP algorithm. The algorithm is an iterative procedure of updating the fragment of the semantic network on personal meanings (updating current situation). Planning process is a forward process starting from the initial situation and ending with the goal situation. Iterations continue until a prediction matrix of the current situation will not be included into the prediction matrix of the initial situation. 

Each MAP iteration consists of three basic steps. At the first step, M-step, algorithm selects such significances of signs that belong to the situation which allow it's role assignment i.e. concretization in the conditions of the situation. At the next step, A-step, algorithm generates specified actions based on selected significances and rates these actions within personal meanings of corresponding signs formed in previous planning experience. Various heuristics can be applied during the A-step. Each heuristic is the inner property of the cognitive process of the agent. At the same time it is mediated by some sign and can be modified (as any sign) during the learning process, which demonstrates that sign approach has wide opportunities for metacognitive modeling.

At the last step, P-step, algorithm integrates conditions of selected actions for generation of the next situation (previous situation in the term of the forward planning). If the start situation is reached algorithm can either stop the planning process with the resultant plan being a sequence of images of constructed actions, or continue by further expansion of images of plan actions (plan specification). In the first case, planning phase is over and plan realization is launched by activating the features composing the plan actions (including activation of path planning operations). The second case can be interpreted as an abstraction level decrease and translation of sub-symbol level into symbol level by considering of sub-plans. Initial situation of the sub-plans is the set of features belonging to conditions of corresponding action and final situation is the set of features belonging to effects of corresponding action. Choosing between plan realization and plan specification is the task of the metacognitive regulation.

\subsection{Path planning}
\label{sect:path}

We follow a well established in AI community approach of formulating path planning problem as a graph search task \cite{ferguson2005guide}. Under this approach two main subtasks are to be decided: first, graph model of the environment (and possibly of agent's dynamic and kinematic constraints) should be constructed; second, search for a path on that graph should be performed resulting in finding feasible path (or in indicating that no path exists). We use 2D regular grid to represent the environment as it is simple yet informative model and can be handled effectively by almost any known search algorithm \cite{yap2002grid}. To construct a grid out of the initial environment's description one need to fix the grid cell size. Fixing this size is very important and can affect path planning in many ways. For example when the cell size is relatively small the grid model of a large workspace contains way too many cells and searching for a path becomes computationally burdensome (although the resultant paths are more precise). If the size is too big than it is easier (in computational sense) to search for a path but the later becomes more ambiguous. Also in that case some obstacles can merge within the graph model in a way that it is impossible to find a graph path although the path in a workspace does exist. In our work we set the grid cell to be equal to the goal area which is described on sub-symbolic level as a square. We also assume that the cell's side size is bigger than $2R$ which means each cell can fully accommodate the agent. We use center based grid notation so agent's positions are tied to the centers of the cells. Horizontal, vertical and diagonal moves are allowed between any two adjacent traversable cells except diagonal moves along blocked cells (this is done to avoid hitting a sideway obstacle when moving).

After the grid is constructed and the agent's transition rules are fixed search for a path, which is a sequence of traversable cells ending with the goal cell, is done. We use two well-known heuristic search algorithms to accomplish that task: A* \cite{hart1968formal} which is the golden standard in AI and it's  modification called Jump Point Search \cite{harabor2011online} The latter is known to find optimal solutions, e.g. the shortest paths, using much less computational effort. The reason why we use both JPS and A* will be explained further. For both algorithms we suggest using octile distance as the heuristic to guide the search, although any other admissible heuristic can be used.

When the sub-symbolic planner is invoked, grid is constructed and goal is set, JPS algorithm is used to search for a path. In case path is found success is reported to the symbolic planner. More interesting case is when JPS returns \textit{failure} to find the path. As JPS is sound and complete it means that no grid path between fixed start cell and goal cell exists which, in turn, means that there is an obstacle (at least one) blocking the path. The task of the path planner now is to identify that obstacle and in case it's destroyable pass this information to the symbolic planner. The latter in that case will be able to form the sub-goal of approaching the obstacle and destroying it. Suppose we know the cell which is adjacent to the blocking obstacle --- seed cell. In that case we invoke simple bug-like wall-following algorithm to mark the contour of that obstacle, e.g. to identify all grid cells adjacent to untraversable cells forming the blocking obstacle. We name such set of cells as \textit{CONTOUR}. And it is \textit{CONTOUR} now that can be passed to symbolic planner as a result (with translation to agent-centered polar coordinate frame if needed). So the task now is to find one of the seed cell.

To identify the seed cell obstacle we use A* algorithm. The reason JPS is not suitable for that task is that JPS skips large portions of space while performing the search and thus achieving higher rates of efficiency. Occasionally it happens so that the explored space, so called CLOSED set, doesn't contain any seed cell. A* on the contrary explores all the free space available and it's \textit{CLOSED} list obligatory contains all the seed cells. To identify them we iterate through the \textit{CLOSED} set in order to find the cell with the lowest value of heuristic function (h-value). This is the cell which is the closest (in metric sense) to the goal which means it is obligatory adjacent to the blocking obstacle. So now we invoke wall-following algorithm to form the \textit{CONTOUR} set and pass the latter to symbolic planner. Symbolic planer now triggers the re-planning procedure which may (and actually will) alter the current action-plan and include obstacle destruction action into it

\section{Experiments}
\label{sect:experiments}
Behavior planning algorithm, MAP, was implemented as Python program library. Available actions, start and goal situations were described with a PDDL language. To map predicates and actions used in PDDL translation function that generates corresponding signs was implemented. Start and goal situations were mapped to the initial and final fragments of the network on personal meanings of signs. Each iteration of the MAP algorithm generated new fragment of the network and defined the next situation.

\begin{figure}[tb]
	\begin{centering}
		\includegraphics[width=\textwidth]{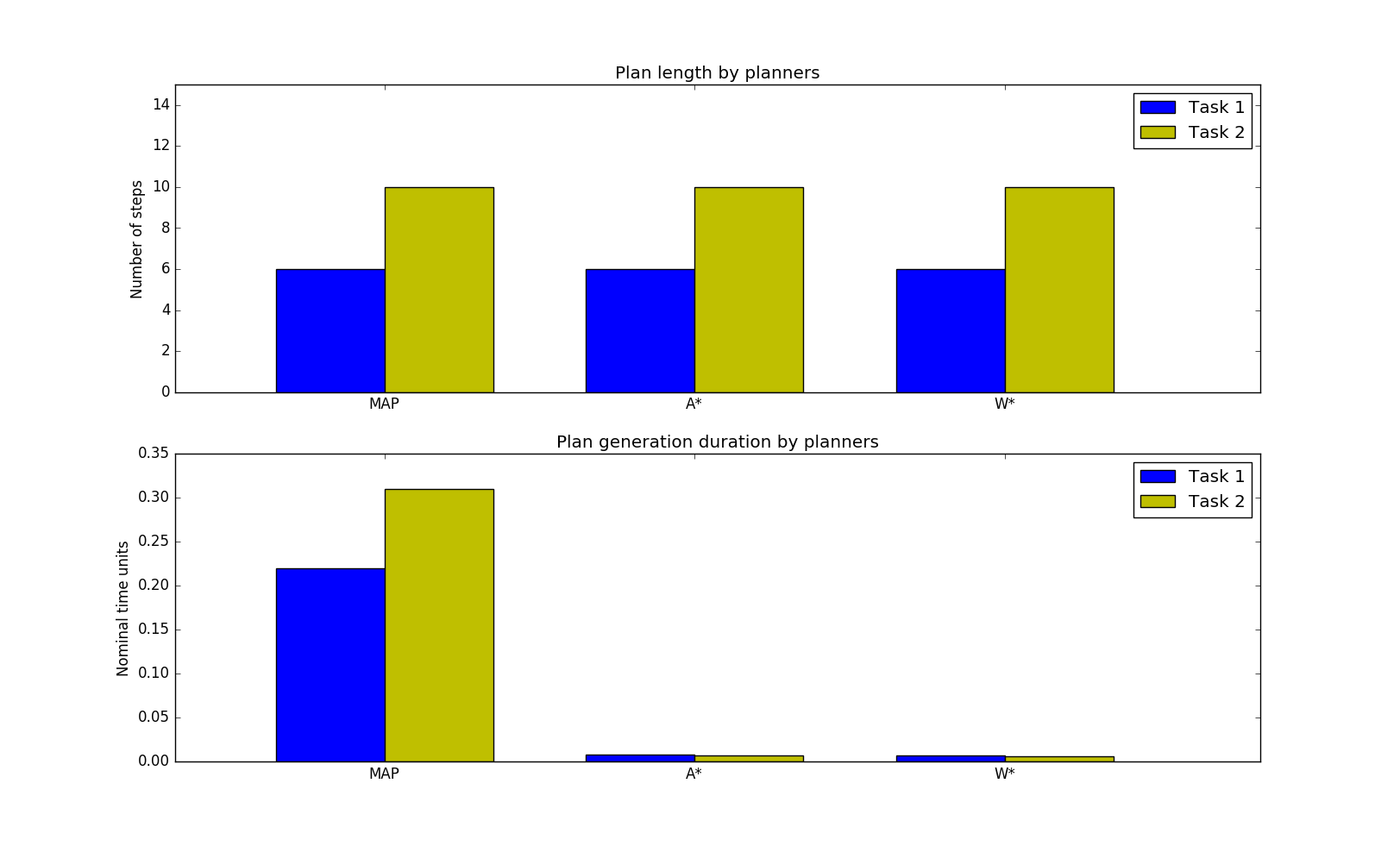}
		\caption{Comparative results of STRIPS task planning for behavior planners}
		\label{fig:beh_exper}
	\end{centering}
\end{figure}

Preliminary experiments were carried out on the well-known ``Block world'' task. Obtained path lengths and runtimes are presented on the Figure~\ref{fig:beh_exper} for MAP planner and planners using standard heuristics (A* and W*) to implement a graph search procedure. As one can see performance of the MAP planner is significantly lower due to large overhead on the initial process of semiotic network generation. At the same time lengths of constructed plans are same for all planners. We believe that time overheads caused by the initial preparation and the supposed learning process will become irrelevant when solving more complex smart relocation tasks.

Path planning algorithms were evaluated on $500\times500$ grids modeling fragments of city environment. Real data from Openstreetmaps database (openstreetmaps.org) was used to construct the grids. On each grid start and goal cells were chosen on the opposite sides (in a way that a distance between them exceed 400) and a goal-blocking obstacle was added. Blockage of the goal cell was organizing in different ways. First, a solid cell line dividing the grid into two almost equal areas --- one containing the goal cell and one containing the start cell --- was added. Second, quarter of a grid containing goal cell was blocked by two perpendicular solid cell lines. Finally, goal cell was blocked by square-shaped obstacle with the side of the square equal to 100, 50 and 10 cells. Further, we refer to these different types of blockage as \textit{Half}, \textit{Quarter}, \textit{Bagel-100}, \textit{Bagel-50} and \textit{Bagel-10}. Each experiment was run in accordance with the presented path-planning loop. First, JPS algorithm was invoked. If it returned \textit{failure} to find a path (which was always a case) then modified A* capable of identifying blocking obstacle contour was launched.

\begin{figure}[tb]
	\begin{centering}
		\includegraphics[width=\textwidth]{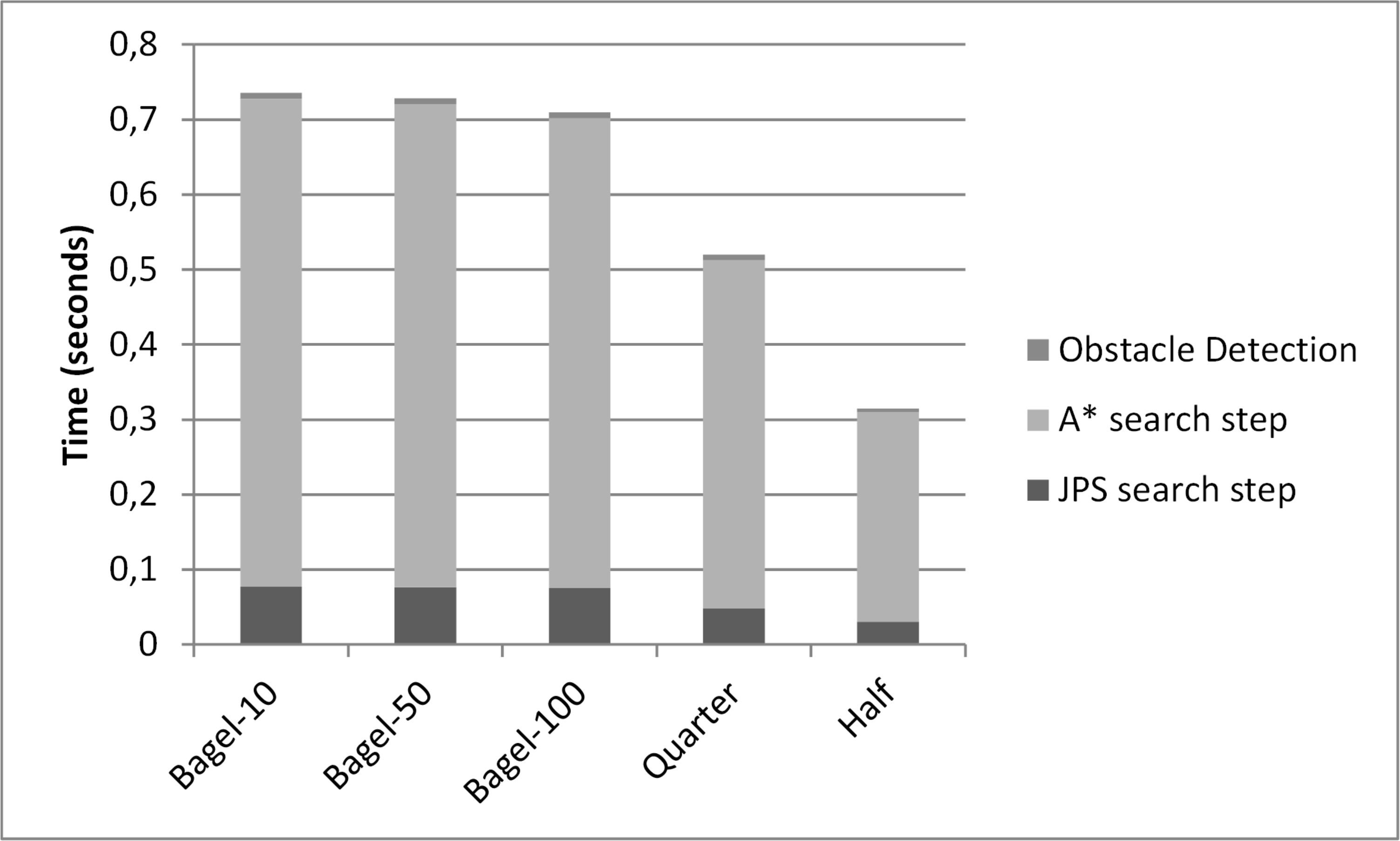}
		\caption{Path planning runtimes}
		\label{fig:path_exper}
	\end{centering}
\end{figure}

Results of the experiments, e.g. runtime, are presented in Figure~\ref{fig:path_exper}. Depending on the type of blockage it took 30-77ms of JPS runtime to claim that path doesn't exist. Then it took 279-650 ms to accomplish A* search which explored all the free space. Finally it took 5-8 ms to identify the seed cell and the blocking obstacle. 

As one can see, the type of blockage significantly affects all phases of the path planning process. The worst search performance is observed for \textit{Bagel-10} grids, which is not surprising --- in this case almost 99,9\% of the free grid cells should be visited to assure the fact path not exists (99\% for \textit{Bagel-50}, 96\% for \textit{Bagel-100}, 75\% for \textit{Quarter} and 50\% for \textit{Half}). Obstacle detection procedure works twice faster for Half-blocked grid than for Bagels. We believe this happens because the most time-consuming step of this procedure is identifying the seed cell, which is done via iterating over the set of all visited cells (which is bigger for Bagel grids). Normalized results look as follows: no matter which type obstacle blockage is present, JPS step takes 10\% of time, A* step 87\%, and obstacle detection step 3\%. Finally, we compared the achieved runtimes with the runtime of JPS launched on the same grids but with no blocking obstacles present: 602ms vs 6,3ms on average. As one can see smart path planning (based on the methods we suggest) is two orders of magnitude more resource intensive than simply estimating the fact that no path can be found on a given grid. We also believe that more efficient approaches for path planning with blocking obstacle detection can be proposed and it is an appealing direction of future work.
\section{Summary}
\label{sect:summary}

We have presented a novel method of behavior planning for modern cognitive architectures suitable for robot control. To demonstrate this method a special class of robot navigation and control tasks, e.g. smart relocation tasks, was considered. The method itself is based on the hierarchical two-level model of abstraction and knowledge representation, e.g. symbolic and subsymbolic. On the symbolic level sign world model is used for knowledge representation and hierarchical planning algorithm, MAP, is utilized for planning. On the subsymbolic level the task of path planning is considered and solved as a graph search problem. Interaction between both planners was examined and inter-level interfaces and feedback loops were described. Preliminary experimental results were presented. We see main strands of future work in conducting experiments on real robotics systems (with real navigation actions) and including learning phase for training metacognitive rules.

\textbf{Acknowledgments}. This work was supported by the Russian Science
Foundation (Project No. 16-11-00048).
%
\label{sect:bib}
\bibliographystyle{plain}
\bibliography{behpath}

\begin{thebibliography}{10}

\bibitem{DellaPenna2012}
G.~{Della Penna}, D.~Magazzeni, and F.~Mercorio.
\newblock {A universal planning system for hybrid domains}.
\newblock {\em Applied Intelligence}, 36(4):932--959, 2012.

\bibitem{Emelyanov2016}
S.~Emel'yanov, D.~Makarov, A.~I. Panov, and K.~Yakovlev.
\newblock {Multilayer cognitive architecture for UAV control}.
\newblock {\em Cognitive Systems Research}, 39:58--72, sep 2016.

\bibitem{ferguson2005guide}
D.~Ferguson, M.~Likhachev, and A.~Stentz.
\newblock A guide to heuristic-based path planning.
\newblock In {\em Proceedings of the international workshop on planning under
  uncertainty for autonomous systems, international conference on automated
  planning and scheduling (ICAPS)}, pages 9--18, 2005.

\bibitem{Fikes1971}
R.~E. Fikes and N.~J. Nilsson.
\newblock {STRIPS: A new approach to the application of theorem proving to
  problem solving}.
\newblock {\em Artificial Intelligence}, 2(3-4):189--208, 1971.

\bibitem{Hanford2011}
S.~D. Hanford.
\newblock {\em {A cognitive robotic system based on the SOAR cognitive
  architecture for mobile robot navigation, search and mapping mission}}.
\newblock PhD thesis, The Pennsylvania State University, 2011.

\bibitem{harabor2011online}
D.~D. Harabor, A.~Grastien, et~al.
\newblock Online graph pruning for pathfinding on grid maps.
\newblock In {\em Proceedings of the Twenty-Fifth AAAI Conference on Artificial
  Intelligence (AAAI-11)}, 2011.

\bibitem{Harnad1990}
S.~Harnad.
\newblock {Symbol Grounding Problem}.
\newblock {\em Physica}, 42:335--346, 1990.

\bibitem{hart1968formal}
P.~E. Hart, N.~J. Nilsson, and B.~Raphael.
\newblock A formal basis for the heuristic determination of minimum cost paths.
\newblock {\em IEEE Transactions on Systems Science and Cybernetics},
  4(2):100--107, 1968.

\bibitem{Herskovits1997}
A.~Herskovits.
\newblock {Language, Spatial Cognition, and Vision}.
\newblock In O.~Stock, editor, {\em Spatial and Temporal Reasoning}, pages
  155--202. Springer, 1997.

\bibitem{Hoffmann2001}
J.~Hoffmann and B.~Nebel.
\newblock {The FF Planning System: Fast Plan Generation Through Heuristic
  Search}.
\newblock {\em Journal of Artificial Intelligence Research}, 14:253--302, 2001.

\bibitem{Kuipers2000}
B.~Kuipers.
\newblock {Spatial semantic hierarchy}.
\newblock {\em Artificial Intelligence}, 119(1):191--233, 2000.

\bibitem{Leontyev2009}
A.~N. Leontyev.
\newblock {\em {The Development of Mind}}.
\newblock Erythros Press and Media, Kettering, 2009.

\bibitem{Lieto2014}
A.~Lieto.
\newblock {A Computational Framework for Concept Representation in Cognitive
  Systems and Architectures: Concepts as Heterogeneous Proxytypes}.
\newblock {\em Procedia Computer Science}, 41:6--14, 2014.

\bibitem{Lieto2016}
A.~Lieto and D.~P. Radicioni.
\newblock {From human to artificial cognition and back: New perspectives on
  cognitively inspired AI systems}.
\newblock {\em Cognitive Systems Research}, 39:1--3, 2016.

\bibitem{Machery2011}
E.~Machery.
\newblock {\em {Doing without Concepts}}.
\newblock Oxford University Press, Oxford, 2011.

\bibitem{Osipov2014b}
G.~S. Osipov, A.~I. Panov, and N.~V. Chudova.
\newblock {Behavior control as a function of consciousness. I. World model and
  goal setting}.
\newblock {\em Journal of Computer and Systems Sciences International},
  53(4):517--529, 2014.

\bibitem{Osipov2015e}
G.~S. Osipov, A.~I. Panov, and N.~V. Chudova.
\newblock {Behavior Control as a Function of Consciousness. II. Synthesis of a
  Behavior Plan}.
\newblock {\em Journal of Computer and Systems Sciences International},
  54(6):882--896, 2015.

\bibitem{Panov2016a}
A.~I. Panov and K.~S. Yakovlev.
\newblock {Behavior and path planning for the coalition of cognitive robots in
  smart relocation tasks}.
\newblock In Jong-Hwan Kim, Fakhri Karray, Jun Jo, Peter Sincak, and Hyun
  Myung, editors, {\em Robot Intelligence Technology and Applications 4},
  Advances in Intelligent Systems and Computing, page (In Press). 2016.

\bibitem{Osipov1997b}
D.~A. Pospelov and G.~S. Osipov.
\newblock {Knowledge in semiotic models}.
\newblock In {\em Proceedings of the Second Workshop on Applied Semiotics,
  Seventh International Conference on Artificial Intelligence and
  Information-Control Systems of Robots (AIICSR'97)}, pages 1--12, Bratislava,
  1997.

\bibitem{Richter2010}
S.~Richter and M.~Westphal.
\newblock {The LAMA planner: Guiding cost-based anytime planning with
  landmarks}.
\newblock {\em Journal of Artificial Intelligence Research}, 39:127--177, 2010.

\bibitem{Schank1972}
R.~C. Schank.
\newblock {Conceptual dependency: A theory of natural language understanding}.
\newblock {\em Cognitive Psychology}, 3(4):552--631, 1972.

\bibitem{Skrynnik2016}
A.~Skrynnik, A.~Petrov, and A.~I. Panov.
\newblock {Hierarchical temporal memory implementation with explicit states
  extraction}.
\newblock In Alexei~V. Samsonovich, Valentin~V. Klimov, and Galina~V. Rybina,
  editors, {\em Biologically Inspired Cognitive Architectures (BICA) for Young
  Scientists}, Advances in Intelligent Systems and Computing, pages 219--225.
  Springer International Publishing, 2016.

\bibitem{Steels2012}
L.~Steels and M.~Hild, editors.
\newblock {\em {Language Grounding in Robots}}.
\newblock Springer US, 2012.

\bibitem{Sun2012a}
R.~Sun and S.~H{\'{e}}lie.
\newblock {Psychologically realistic cognitive agents: taking human cognition
  seriously}.
\newblock {\em Journal of Experimental {\&} Theoretical Artificial
  Intelligence}, 25(1):65--92, 2012.

\bibitem{Trafton2013}
G.~J. Trafton, L.~M. Hiatt, A.~M. Harrison, F.~P. Tamborello, S.~S. Khemlani,
  and A.~C. Schultz.
\newblock {ACT-R/E: An embodied cognitive architecture for Human-Robot
  Interaction}.
\newblock {\em Journal of Human-Robot Interaction}, 2(1):30--54, 2013.

\bibitem{yap2002grid}
P.~Yap.
\newblock Grid-based path-finding.
\newblock In {\em Advances in Artificial Intelligence}, pages 44--55. Springer,
  2002.

\bibitem{Zadeh2012}
L.~A. Zadeh.
\newblock {\em {Computing with Words}}.
\newblock Studies in Fuzziness and Soft Computing. Springer Berlin Heidelberg,
  2012.

\end{thebibliography}

\end{document}